\documentclass{article}

\usepackage{times}
\usepackage{graphicx} 

\usepackage{natbib}

\usepackage{algorithm}
\usepackage{algorithmic}

\usepackage[draft]{hyperref}

\usepackage[accepted]{icml2016} 

\usepackage{times}
\usepackage{helvet}
\usepackage{courier}
\usepackage{url}
\usepackage{multirow}
\usepackage{pdfpages}

\usepackage[english, status=draft]{fixme}
\fxusetheme{color}

\usepackage{amsmath}
\usepackage{amssymb}

\usepackage{graphicx}
\usepackage{caption}
\usepackage{subcaption}

\usepackage{tikz}
\usetikzlibrary{bayesnet}

\newcommand{\argmin}{\operatornamewithlimits{argmin}}
\newcommand{\gp}{\mathcal{GP}}

\usepackage[acronym]{glossaries}

\newacronym[longplural={Gaussian Processes}]{gp}{GP}{Gaussian Process}
\newacronym{abcd}{ABCD}{Automatic Bayesian Covariance Discovery}
\newacronym{rabcd}{RABCD}{Relational Automatic Bayesian Covariance Discovery}
\newacronym{bic}{BIC}{Bayesian Information Criterion}
\newacronym{mkl}{MKL}{Multiple Kernel Learning}
\newacronym{ckl}{CKL}{Compositional Kernel Learning}
\newacronym{rkl}{RKL}{Relational Kernel Learning}
\newacronym[longplural={Relational Gaussian Processes}]{rgp}{RGP}{Relational Gaussian Process}
\newacronym{srl}{SRL}{Statistical Relational Learning}

\icmltitlerunning{The Automatic Statistician: A Relational Perspective}

\begin{document} 

\twocolumn[
\icmltitle{The Automatic Statistician: A Relational Perspective}

\icmlauthor{Yunseong Hwang}{yunseong@unist.ac.kr}
\icmladdress{Ulsan National Institute of Science and Technology, Ulsan, 44919 Korea}
\icmlauthor{Anh Tong}{anhth@unist.ac.kr}
\icmladdress{Ulsan National Institute of Science and Technology, Ulsan, 44919 Korea}
\icmlauthor{Jaesik Choi}{jaesik@unist.ac.kr}
\icmladdress{Ulsan National Institute of Science and Technology, Ulsan, 44919 Korea}

\icmlkeywords{boring formatting information, machine learning, ICML}

\vskip 0.3in
]

\begin{abstract}
Gaussian Processes (GPs) provide a general and analytically tractable way of modeling complex time-varying, nonparametric functions.
The Automatic Bayesian Covariance Discovery (ABCD) system constructs natural-language description of time-series data by treating unknown time-series data nonparametrically using GP with a composite covariance kernel function.
Unfortunately, learning a composite covariance kernel with a single time-series data set often results in less informative kernel that may not give qualitative, distinctive descriptions of data. 
We address this challenge by proposing two relational kernel learning methods which can model multiple time-series data sets by finding common, shared causes of changes.
We show that the relational kernel learning methods find more accurate models for regression problems on several real-world data sets; US stock data, US house price index data and currency exchange rate data.
\end{abstract}
\glsresetall

\section{Introduction}
\label{sec:intro}

\Glspl*{gp} provide a general and analytically tractable way of capturing complex time-varying, nonparametric functions. The time varying parameters of \glspl*{gp} can be explained as a composition of base kernels such as linear, smoothness or periodicity in that covariance kernels are closed under addition and multiplication. The \gls*{abcd} system \cite{lloyd2014automatic} constructs natural-language description of time-series data by treating unknown time-series data nonparametrically using \glspl*{gp}.

It is important to find data dependencies and the structure in time-series data. \glspl*{gp} represent data in a non-parametric way with a mean function and a covariance kernel function. The covariance kernel function determines correlation patterns between the data points. Therefore, learning a proper kernel is essential to model data points with \glspl*{gp} \cite{rasmussen2006gaussian,DiosanRP07,Bing2010,KlenskeZSH13,lloyd2014gefcom2012}.
Unfortunately, finding an appropriate kernel often requires manual encoding by human experts or reduces to a simple problem estimating parameters of a fixed, predefined kernel structure. 

The covariance kernels of \glspl*{gp} are known to be closed under addition and multiplication \cite{duvenaud4922structure}. Thus, a sequence of complex real-valued variables could be explained by a compositional kernel with base kernels and kernel operations \cite{mkl04,CandelaR05,WilsonA13}. Recently, kernel operation grammars and a framework for an automatic discovery of a compositional kernel have been proposed \cite{lloyd2014automatic}. This framework is flexible and interpretable in that the framework automatically discovers a complex composition of  interpretable base kernels. Once a compositional kernel is found, the individual base components can be construed in a human readable form. This capability of decomposition into multiple components and interpretability shed light on finding shared structure given multiple sets of data.

Finding a shared structure in multiple sequences may reveal the common covariance structures of the sequences beyond the patterns in a single sequence \cite{churelational,XuKT09,GWP}. As a typical example in economics, the multivariate view is central where each variable is normally viewed in the context of relationships to other variables, like exchange rate affecting gross domestic product (GDP).

We propose two Relational Multi-Kernel Learning methods (explained in Section~\ref{sec:RKL}) to find a shared covariance kernel for multiple sets of data sequences. Our algorithm discovers both a shared composite kernel, which explains the common causes of changes in multiple data sets, and individual components (scale factors and distinctive kernels), which explain changes in individual data sets. Since GPs with the learned kernel are non-parametric, we can represent any data which is known to have the same relationship with the training data. 

This paper is organized as follows. Section~\ref{sec:background} introduces \acrfull*{gp} and the \acrfull*{abcd} system. Section~\ref{sec:RKL} presents our main contribution and algorithm, \acrfull*{rabcd}. Section~\ref{sec:related} discusses related work. Section~\ref{sec:exps} shows experimental results with real-world data sets followed by conclusions in Section~\ref{sec:conclusions}.

\begin{figure*}
\centering
\includegraphics[width=\textwidth]{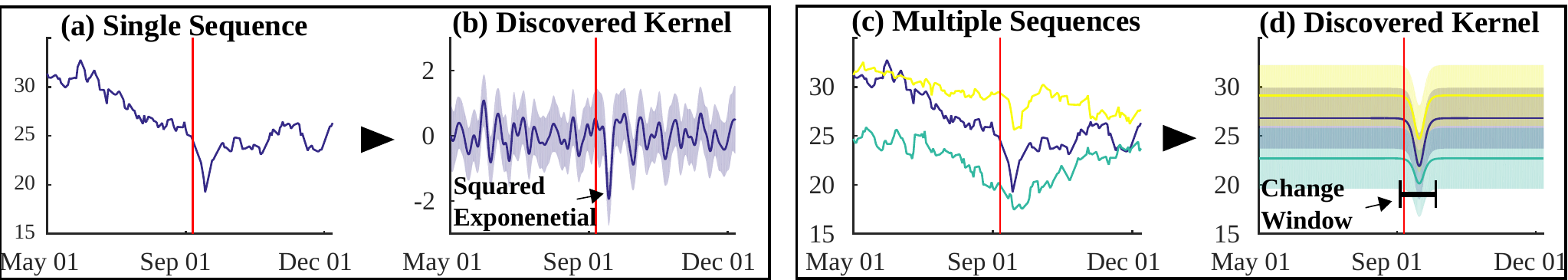}
\caption{This figure shows components learned from \acrshort*{ckl} (left) and \acrshort*{rkl} (right). (a) and (b) represent adjusted closes of GE and learned components by \acrshort*{ckl}, respectively. (c) and (d) represent adjusted closes of 3 selected stocks (from top, yellow:XOM, purple:GE, green:MSFT) and learned (common) components by \acrshort*{rkl}. The red vertical lines indicate the September 11 attacks which occurred in 2001.  \Acrshort*{rkl} finds and explains the sudden drop after 911 by selectively applying a constant kernel around that period. However, \acrshort*{ckl} tries to fit that drop using a (less informative) rapidly varying squared exponential (SE) kernel function. We provide all components learned by \acrshort*{ckl} and \acrshort*{rkl} in the appendix.}
\label{fig:compare}
\end{figure*}

\section{\acrlong*{abcd}}
\label{sec:background}

Here, we briefly explain \acrfullpl*{gp} and then introduce the \acrfull*{abcd} system \cite{lloyd2014automatic}, which is an extension of the \acrfull*{ckl} \cite{duvenaud4922structure}.

\subsection{\acrlong*{gp}}

\Glspl*{gp} are distributions over functions such that any finite set of function evaluations, $(f(x_1), f(x_2), \dotsc, f(x_N))$ form a multivariate Gaussian distribution \cite{rasmussen2006gaussian}. As a multivariate Gaussian distribution is specified by its mean vector $\mathbf{\mu}$ and covariance matrix $\mathbf{\Sigma}$, a \gls*{gp} is specified by its mean function, $\mu(x){=}\mathbb{E}(f(x))$ and covariance kernel function, $k(x,x'){=}\mathrm{Cov}(f(x), f(x'))$. Evaluations of the two functions on a finite set of points correspond to the mean vector and the covariance matrix for the multivariate Gaussian distribution, like $\mathbf{\mu}_i{=}\mu(x_i)$ and $\mathbf{\Sigma}_{ij}{=}k(x_i, x_j)$. 
When a \gls*{gp} is specified, we can calculate the marginal likelihood of given data or can derive predictive distribution for new points given existing data. 

Throughout this paper, we will use the following notations for \glspl*{gp}. If a function or evaluations of the function $f$ are drawn from a \gls*{gp} specified by its mean function $\mu(x)$ and covariance kernel function $k(x,x')$, 
\begin{align*}
f \sim \gp{\left(\mu(x),k(x,x')\right)}
\end{align*}
We may occasionally omit the arguments of the functions, $x$ and $x'$, for simplicity and just say $k$ as a kernel function. For zero-mean \glspl*{gp}, we put 0 in the place of mean function, which means $\mu(x){=}0$. The covariance kernel functions often have its hyperparameters which are free parameters in defining a kernel function. For example, the squared exponential kernel function $k(x,x')=\sigma^2\exp(|x-x'|^2/\ell)$ has two hyperparameters, $\sigma$ and $\ell$. In the following, we will use notation $k(x,x';\theta)$ as a kernel function that has a vector $\theta$ as its hyperparameters.

\subsection{\acrlong*{ckl}}

\Gls*{ckl} constructs and finds richer kernels which are composed of several base kernels and operations. In theory, any positive definite kernels are closed under addition and multiplication \cite{duvenaud4922structure}.
Here, each base kernel expresses each distinctive feature such as linearity or periodicity. The kernel operations include not only addition and multiplication but also the so-called change-point and change-window operation to deal with abrupt structural changes.

\subsubsection{Base Kernels}

Five base kernels are used for making compositional kernels. Each kernel encodes different characteristics of functions, which further enables the generalization of structure and inference given new data.

\begin{tabular}{l | l}
Base Kernels & Encoding Function \\
\hline
White Noise (\textsc{WN}) & Uncorrelated noise \\
Constant (\textsc{C}) & Constant functions \\
Linear (\textsc{LIN}) & Linear functions \\
Squared Exponential (\textsc{SE}) & Smooth functions \\
Periodic (\textsc{PER}) & Periodic functions
\end{tabular}

\subsubsection{Operations}

The first operation is addition which sums multiple kernel functions and makes a new kernel function.
\begin{align*}
k'(x,x') = k_1(x,x') + k_2(x,x').
\end{align*}
This operation works as a superposition of multiple independent covariance functions. In general, if $f_1 \sim \mathcal{GP}(\mu_1, k_1), f_2 \sim \mathcal{GP}(\mu_2, k_2)$ then $f:=f_1+f_2 \sim \mathcal{GP}(\mu_1+\mu_2, k_1+k_2)$.

ABCD assumes zero mean for \glspl*{gp}, since marginalizing over an unknown mean function can be equivalently expressed as a zero-mean \gls*{gp} with a new kernel \cite{lloyd2014automatic}. Under this assumption, the following multiplication operation is also applicable.
\begin{align*}
k'(x,x') = k_1(x,x') \times k_2(x,x').
\end{align*}
In general, if $f_1 \sim \mathcal{GP}(0, k_1), f_2 \sim \mathcal{GP}(0, k_2)$ then $f:=f_1 \times f_2 \sim \mathcal{GP}(0, k_1 \times k_2)$.

The third operation is the change-point (CP) operation. Given two kernel functions, $k_1$ and $k_2$, the new kernel function is represented as follows:
\begin{align*}
\begin{split}
&k'(x,x') = \sigma(x)k_1(x,x')\sigma(x') \\
&\quad+ (1-\sigma(x))k_2(x,x')(1-\sigma(x')) \nonumber
\end{split}
\end{align*}
where $\sigma(x)$ is a sigmoidal function which lies between 0 and 1, and $\ell$ is the change-point.
The change-point operation divides function domain (i.e., time) into two sides and applies different kernel function on each side. To generalize, if $f_1 \sim \mathcal{GP}(\mu_1, k_1), f_2 \sim \mathcal{GP}(\mu_2, k_2)$ then $f:=\sigma(x)f_1 + (1-\sigma(x))f_2 \sim \mathcal{GP}(\sigma(x)\mu_1 + (1-\sigma(x))\mu_2, \sigma(x)k_1\sigma(x') + (1-\sigma(x))k_2(1-\sigma(x')))$.

Finally, the change-window (CW) operation applies the CP operation twice with two different change points $\ell_1$ and $\ell_2$. Given two sigmoidal functions $\sigma_1(x; \ell_1)$ and $\sigma_2(x; \ell_2)$ where $\ell_1 < \ell_2$, the new function will be $f:=\sigma_1(x)f_1(1-\sigma_2(x)) + (1-\sigma_1(x))f_2\sigma_2(x)$, which applies the function $f_1$ to the window $(\ell_1, \ell_2)$. 
A composite kernel expression after the change-window operation will be as follows:
\begin{align*}
&k'(x,x') = \sigma_1(x)(1-\sigma_2(x))k_1(x,x')\sigma_1(x')(1-\sigma_2(x')) \nonumber \\
&\quad+ (1-\sigma_1(x))\sigma_2(x)k_2(x,x')(1-\sigma_1(x'))\sigma_2(x'). \nonumber
\end{align*}

\subsubsection{Search grammar}

\Gls*{abcd} searches a composite kernel based on the search grammar. The search grammar specifies how to develop the current kernel expression by applying the operations with the base kernels. The following rules are examples of typical search grammar:
\begin{align*}
\mathcal{S} &\rightarrow \mathcal{S + B} &&\mathcal{S} \rightarrow \mathcal{S \times B} \\
\mathcal{S} &\rightarrow \mathrm{CP}(\mathcal{S,S}) &&\mathcal{S} \rightarrow \mathrm{CW}(\mathcal{S,S})  \\
\mathcal{S} &\rightarrow \mathcal{B} &&\mathcal{S} \rightarrow \mathrm{C} 
\end{align*}
where $\mathcal{S}$ represents any kernel subexpression, $\mathcal{B}$ and $\mathcal{B}'$ are base kernels. 
For example, supposed that there is a kernel expression $\mathcal{E} = \mathcal{K}_1 + \mathcal{K}_2 + \mathcal{K}_3$ where $\mathcal{K}_i$ is a kernel expression which cannot be separated into summands. Then, kernel subexpression $\mathcal{S}$ can be the summation of a non-empty subset of $\mathcal{K}_i$. If we apply multiplication grammar on a subset ($\mathcal{K}_1 + \mathcal{K}_3$) with a base kernel $\mathcal{B}_1$, the expanded kernel expression is $\mathcal{E}' =\mathcal{K}_2 + (\mathcal{K}_1 + \mathcal{K}_3) \times \mathcal{B}_1$

\subsubsection{Algorithm}

Given data and a maximum search depth, the algorithm gives a compositional kernel $k(x,x';\theta)$. Starting from the \textsc{WN} kernel, the algorithm expands the kernel expression based on the search grammar, optimizes hyperparameters for the expanded kernels, evaluates those kernels given the data and selects the best one among them. This procedure repeats. The next iterative procedure starts with the best composite kernel selected in the previous iteration. The conjugate gradient method is used when optimizing hyperparameters. \Gls{bic}~\cite{schwarz1978estimating} is used for the model evaluation. The \gls{bic} of model $\mathcal{M}$ with $|\mathcal{M}|$ number of free parameters and data $\mathcal{D}$ with $|\mathcal{D}|$ number of data points is:
\begin{equation}
\label{eq:bic}
\textsc{BIC}(\mathcal{M}) = -2\log p(\mathcal{D}|\mathcal{M}) + |\mathcal{M}| \log |\mathcal{D}|.
\end{equation}
The iteration continues until the specified maximum search depth is reached. During the iteration, the algorithm keeps the best model for the output.

Here is an example of how this algorithm works. Suppose that we start from the \textsc{WN} kernel. Then, we apply operators with some base kernels as described in the expansion grammar, such as $\textsc{WN} \rightarrow \textsc{WN} + \textsc{SE}$, $\textsc{WN} \rightarrow \textsc{WN} + \textsc{LIN}$ and so on. With those expanded kernels, the algorithm optimizes hyperparameters of each expanded kernel. Now we have all the optimized kernels. Finally we compare the optimized ones and select the best kernel in terms of the \gls{bic}. This procedure repeats until we meet a certain depth of search. As an example, suppose that the best kernel selected in the previous step was $\textsc{WN} + \textsc{SE}$. Then we apply kernel expansion again like $\textsc{WN} + \textsc{SE} \rightarrow \textsc{WN} + \textsc{SE} \times \textsc{LIN}$ and so on. Then we optimize the hyperparameters, select the best kernel, and then proceed to the next step.

\section{Relational Multi-Kernel Learning}
\label{sec:RKL}

\Gls*{rkl} combines the \gls*{abcd} system with \gls*{srl}. Given multiple sets of time-series data, RKL finds a composite kernel for GPs, which can describe the shared structure of the sets. We first introduce RKL model with only a shared kernel, then presents semi-RKL model with shared and individual kernels. 

\subsection{Relational Kernel Learning}

RKL aims to find a model $\mathcal{M}$ that explains $M$ multiple data sets, $\mathcal{D} = \{d_1, d_2, \dotsc, d_M\}$ well. To do so, we find a model that maximizes the likelihood, $p(\mathcal{D}|\mathcal{M}) = p(d_1, d_2, \dotsc, d_M|\mathcal{M})$. Here, $\mathcal{M}$ represents GP models. We assume the conditional independence of the likelihoods of each time-series data $d_i$. Hence, the log likelihood of the whole data is the sum of the log likelihoods of individual time-series:
\begin{align}
\begin{split}
\log p(d_1, \dotsc, d_M|\mathcal{M}) &= \log \prod\nolimits_i p(d_i|\mathcal{M}) \\
&= \sum\nolimits_i \log p(d_i|\mathcal{M})
\end{split}
\end{align}
This notation makes us to consider each data separately and utilization of existing optimization technique and model evaluation form used for a single data set.

\begin{figure}
\centering
\begin{tikzpicture}[scale=0.9]
\node at (-2,0) (gp) {$\gp{}$};
\node at (-3,-2) [latent] (fi) {$f_i$};
\node at (-1,-2) [latent] (fj) {$f_j$};
\node at (-2,-1) [latent] (ff) {$f$};
\node at (-3,-3) [obs] (yi) {$y_i$};
\node at (-1,-3) [obs] (yj) {$y_j$};
  
  \node at (-2,-2.5) [latent] (r) {$R_{ij}$};
\plate {pp} {(fi)(fj)(r)(yi)(yj)} {$i,j \in 1 \dotsc N$};
\edge {gp} {ff};
\edge {ff} {fi};
\edge {ff} {fj};
\edge {fi} {r};
\edge {fj} {r};
\edge {fi} {yi};
\edge {fj} {yj};
\node at (-2,-5) (rgp) {(a) \acrshortpl*{rgp}};
\node at (2,0) (gp) {$\gp{}$};
\node at (2,-1) [latent] (f) {$f$};
\node at (2,-2) [latent] (fk) {$f_i$};
\node at (2,-3) [obs] (y) {$y_i$};
\plate {p} {(fk)(y)} {$i \in 1 \dotsc N$};
\plate {m} {(f)(p)} {$M$};
\edge {gp} {f};
\edge {f} {fk};
\edge {fk} {y};
\node at (1.8,-5) (rkl) {(b) \acrshort*{rkl}};
\end{tikzpicture}
\caption{Graphical representation of (a) \acrshortpl*{rgp}~\cite{churelational} and (b) \acrshort*{rkl}. Here $f$ denotes latent function. The subscripts of $f_i$ and $f_j$ denote sampled latent values for each point, as in $f_i = f(x_i)$. $y_i$ and $y_j$ are observed values for each point. $N$ is the number of data points for each data set. $M$ is the number of different data sets.}
\label{fig:graphical}
\end{figure}
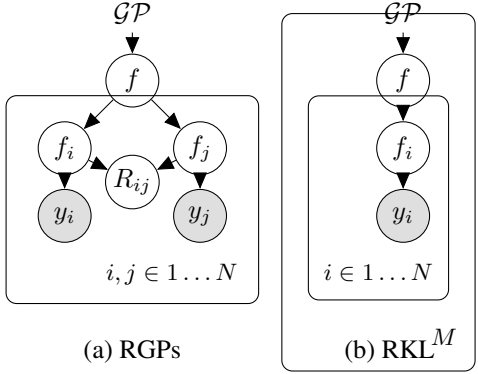

We assume that there is a single, shared factor that determines the covariance pattern for multiple data sets within the same domain, e.g., stock values of companies in a certain  industrial sector. This acts as tying different factors that determine the covariance pattern of each data set as one single factor. By finding the factor, we achieve the shared compositional covariance kernel. Finding the shared factor in GPs is reduced to finding a shared covariance kernel function.

\subsubsection{Description}

Defining a GP requires mean function $\mu(x)$ and covariance function $k(x,x')$. The mean function is set to be a constant function which gives 0. The covariance function is divided into two parts, functional part $k$ and its parameters $\theta$. 
Here we say the functional part $k$ is the function structure but the values of its hyperparameters are not specified yet. $k$ is from a context-free language set yielded from context-free grammar $G$, which defines the search method while expanding the search tree in the kernel search algorithm. Here $G$ contains base kernels like \textsc{WN} and operators like $+$ and $\times$.
After functional part $k$ is decided, the hyperparameter vector $\theta$ is decided. In addition, we have parameter vector $\sigma$ where its length is twice of the number of datasets. Each $b_j$ (additive scaling factor) and $v_j$ (multiplicative scaling factor) in $\sigma=[b_1,v_1,\dotsc,b_M,v_M]$ is added and multiplied to the kernel function $k(x,x';\theta)$ before defining a \gls{gp} prior for each data, i.e. $k' = b_j^2 + v_j^2 \times k(x,x';\theta)$ for each $j$th data set. This parameters normalize of scale variance to deal with the data sets with different scale of bias and variance in target value $y$. Summarizing all the process gives the following model.
\begin{align}
\begin{split}
&k \leftarrow G, \qquad \gp{} \leftarrow k, \theta, \sigma \\
&f \leftarrow \gp{}, \qquad \mathbf{y} \leftarrow f, \mathbf{X}
\end{split}
\end{align}
Here $f$ is the latent function from \gls*{gp} prior. $\mathbf{y}$ and $\mathbf{X}$ are $N \times 1$ and $N \times D$ matrices where $N$ is the number of data points for each data set and $D$ is the input dimension. For $y_i$ which is the $i$th element of column vector $\mathbf{y}$, and $\mathbf{x}_i$ which is the $i$th row vector of matrix $\mathbf{X}$, $y_i = f(\mathbf{x}_i)$ holds.

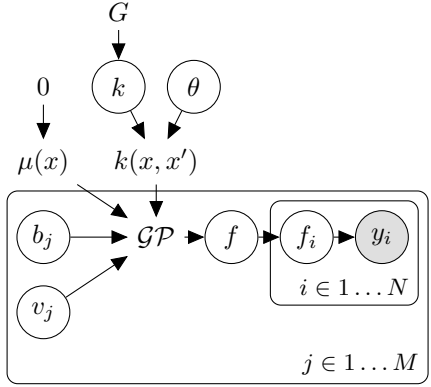
\begin{figure}
\centering
\begin{tikzpicture}
\node at (0,1) (gp) {$\mathcal{GP}$};
\node at (1,1) [latent] (f) {$f$};
\node at (2,1) [latent] (fk) {$f_i$};
\node at (3,1) [obs] (y) {$y_i$};
\node at (-1.5,1) [latent] (sfa) {$b_j$};
\node at (-1.5,0) [latent] (sfm) {$v_j$};
\plate {p} {(fk)(y)} {$i \in 1 \dotsc N$};
\plate {m} {(f)(p)(gp)(sfa)(sfm)} {$j \in 1 \dotsc M$};
\edge {gp} {f};
\edge {f} {fk};
\edge {fk} {y};
\edge {sfa} {gp};
\edge {sfm} {gp};
\node at (-1.5,2) (mux) {$\mu(x)$};
\node at (-1.5,3) (zero) {$0$};
\node at (0,2) (kxx) {$k(x,x')$};
\node at (-0.5,3) [latent] (k) {$k$};
\node at (0.5,3) [latent] (theta) {$\theta$};
\edge {mux} {gp};
\edge {zero} {mux};
\edge {kxx} {gp};
\edge {k} {kxx};
\edge {theta} {kxx};
\node at (-0.5,4) (g) {$G$};
\edge {g} {k};
\end{tikzpicture}
\caption{Graphical representation of \gls*{rkl} model with more steps in generating \gls*{gp} prior. $k$ defines the kernel function's structure but not the values of its hyperparameters. $\theta$ completes the function by deciding the function's hyperparameter values. $b_j$ and $v_j$ are added and multiplied to $k(x,x')$ for each \gls*{gp} prior. This works as normalization of shift and scale variance of each data set.}
\label{fig:graphicaldetail}
\end{figure}

Figure~\ref{fig:graphicaldetail} shows the overall structure of the model. The GP prior is specified for each data set, from 1 to $M$. A mean function (which is zero here) and a kernel function specify a GP prior. The covariance kernel function is specified with the kernel structure and hyperparameters. Additive and multiplicative 
scaling factors, $b_j$ and $v_j$ for each data set, are added and multiplied to the kernel function and specifies a GP prior. 
The GP prior for each data set gives latent function $f$. From that latent function, we get latent values $f_i$ for each data point. Finally $y_i$ is generated from the latent value $f_i$. So for each data set, we have the following distribution:
\begin{gather}
K_{i,j} = b^2 + v^2 \times k(\mathbf{x}_i, \mathbf{x}_j;\theta) \\
\mathbf{y} \sim \mathcal{N}(0, K)
\end{gather}
Here $\mathcal{N}(0,K)$ denotes a multivariate normal distribution with 0 mean vector and covariance matrix $K$.

\subsubsection{Learning}

Algorithm~\ref{alg:model} presents a learning procedure to find a \gls*{rkl} model. This algorithm finds a single covariance kernel, shared by multiple data sets. To explain each line in detail, line~1 limits depth of search tree. Line~2 expands the given kernel $k$ to multiple candidate kernels based on the expansion grammar $G$ giving the set of expanded kernels $K$. For each expanded kernel, in line~4, algorithm optimizes hyperparameters and scaling factors of the kernel, that minimizes negative log-likelihood on the whole multiple data sets $\mathcal{D}$. In the optimization process, conjugate gradient descent algorithm is used. After optimizing all the parameters for every candidate kernels, in line~6, it selects the best kernel among those candidates that minimizes \gls*{bic}. At the end, in line~8, it returns the best kernel.

The main difference is that we calculate the negative log marginal likelihood of the whole data as a summation of negative log marginal likelihood of each data set. This is possible because our \gls*{rkl} model shares a covariance kernel function through multiple data sets. This does not specify any correlation of variables across the different data sets. In other words, if we construct a single covariance matrix for all the data, we get a covariance matrix that has a block diagonal form. Thus we can factorize the likelihood $p(\mathcal{D}|\mathcal{M})$, where $\mathcal{M}$ is a model and $D$ is a set of data, into a product of individual likelihoods $\prod_{d \in \mathcal{D}}p(d|\mathcal{M})$ as those individual likelihoods are independent. As we are dealing with the log likelihood, this can be represented as  $\sum_{d \in \mathcal{D}}\log{p(d|\mathcal{M})}$. 

\begin{algorithm}[t!]
\caption{\acrlong*{rkl}}\label{alg:model}
\label{Rkl}
\begin{algorithmic}[1]
\REQUIRE initial kernel $k$, initial hyperparameters $\theta$, initial scaling factors $\sigma$, multiple data sets $\mathcal{D}=\{d_1,\dotsc,d_M\}$, expansion grammar $G$, maximum depth of search $s$ 
\FOR{$i \in 0 \dotsc s$} \label{line:maxdepth}
\STATE $K \gets$ expand($k,G$) \label{line:expand}
\FOR{$k \in K$}
\STATE $k(\theta,\sigma) \gets \argmin_{(\theta,\sigma)}{-\log p(\mathcal{D}|k)}$ \label{line:hypopt}
\ENDFOR
\STATE $k \gets \argmin_{k \in K}{\mathrm{BIC}(k,\mathcal{D}})$ \label{line:modelsel}
\ENDFOR
\STATE \textbf{return} $k$
\end{algorithmic}
\end{algorithm}


\subsection{Semi-Relational Kernel Learning}
The assumption made by the RKL model is rather too strong to accommodate variations of individual data sequences. To handle this issue, we propose, Semi-Relational Kernel Learning which loosens RKL's constraint by considering tow parts of structure including a shared structure and an individualized structure.
\subsubsection{Description}
SRKL aims to learn a set of kernels 
$$\mathbf{K} = \{K_j = K_S + K_{d_j} | {d_j} \in \mathcal{D}, j = 1, \dotsc , M\},$$
where $K_S$ is the shared kernel among M sequences, $K_{d_j}$ is the distinctive kernel for $j$-th sequence. When we describe time series as types of trees, the shared kernel represents the common shape of trunk shared among those trees. The distinctive kernel interprets the small branches and leaves.

Ideally, the search grammar can be performed to discover one shared kernel and $M$ distinctive kernels. The search space explodes in term of complexity $\mathcal{O}(n^{M+1})$ where $n$ is the number of possible kernels on each search grammar tree for every depth. To reduce this intensive search, the distinctive kernel for each time series does not follow the extensive search grammar but is fixed by using the spectral mixture (SM) kernel ~\cite{WilsonA13} 
$$k(\tau) = \sum_{q=1}^Qw_q\prod\exp\{-2\pi^2\tau_p^2v_q^{(p)}\} \cos (2\pi\tau_p\mu_q^{(p)})$$
where $Q$ is the number of components, $\tau = x - x'$ is a $P$ dimensional vector. SM is chosen because its expressiveness and ability to approximate a wide range of covariance kernels. 
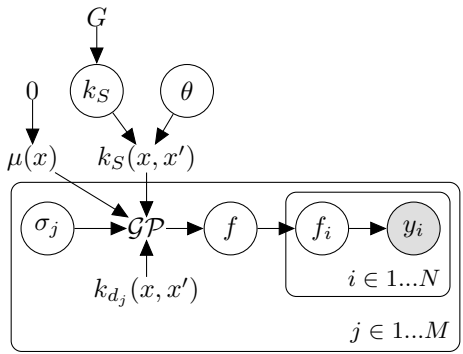
\begin{figure}[t]
  \begin{center}
    \begin{tikzpicture}

  \node[obs]                               	(y) {$y_i$};
  \node[latent, left=0.5cm of y] 				(fi) {$f_i$};
  \node[latent, left= 0.5cm of fi]  			(f) {$f$};
  \node[const, left=0.5cm of f]            	(gp) {$\mathcal{GP}$};
  \node[latent, left=0.7cm of gp]				(s) {$\sigma_j$};
  \node[const, below=0.5cm of gp]				(kj) {$k_{d_j}(x,x')$};
  \node[const, above=0.6cm of gp]				(ks) {$k_S(x,x')$};
  \node[const, left=0.5cm of ks]				(mu) {$\mu(x)$};
  \node[const, above=0.6cm of mu]				(zero) {$0$};
  \node[latent, right=0.4cm of zero]				(k_s) {$k_S$};
  \node[latent, right =1.6cm of zero]				(theta) {$\theta$};
   \node[const, above=0.5cm of k_s]				(g) {$G$};


  \edge {fi} {y} ;
  \edge {f} {fi} ;
  \edge {gp} {f} ;
  \edge {kj} {gp};
  \edge {ks} {gp};
  \edge {mu} {gp};
  \edge {zero} {mu};
  \edge {g}{k_s};
  \edge {k_s}{ks};
  \edge {theta}{ks};

  \edge {s} {gp};

  \plate {fiyi} {(y) (fi)} {$i \in 1 ... N$};
  \plate {M} {(fiyi) (f) (gp) (kj) (s)} {$j \in 1 ... M$}

\end{tikzpicture}
  \end{center}
  \caption{Graphical representation of SRKL model. Comparing to RKL, SRKL elaborates the GP kernel structure with an additional distinctive kernel for each data sequence.}
\end{figure}

\subsubsection{Learning}

\begin{algorithm}[t!]
\caption{Semi-Relational Kernel Learning}
\label{srkl}
\begin{algorithmic}[1]
\REQUIRE data $\mathcal{D}=\{d_1,\dotsc,d_M\}$, grammar $G$, maximum depth of search $s$
\STATE Initialize $\mathcal{K} \gets \emptyset$
\FOR{$i \in 0 \dotsc s$}
\STATE $K_S \gets $ expand(G)
\STATE $\Theta \gets \emptyset$
\FOR{$k_S \in K_S$}
\STATE Initialize  $\theta^0 \gets (\theta^0_S, \theta^0_1,\dotsc, \theta^0_M, \sigma^0_1,\dotsc, \sigma^0_M)$
\STATE $k_j(\theta^0) \gets k_S(\theta_S^0, \sigma^0_j) + k_{d_j}(\theta^0_j), j = 1 \dotsc M$
\STATE $\theta^* \gets \argmin \sum_{j = 1}^M-\log p(\mathcal{D} | k_j(\theta))$
\STATE $\Theta \gets \Theta \cup \{(k_S,\theta^*)\}$
\ENDFOR
\STATE $(\hat{k_S}, \hat{\theta}) \gets \argmin_{(k_S, \theta) \in \Theta}$ BIC ($k_S$, $\mathcal{D}$)
\STATE $\mathcal{K} \gets \mathcal{K} \cup {(\hat{k_S}, \hat{\theta}, \hat{\sigma})}$
\ENDFOR
\STATE \textbf{return} $\mathcal{K}$ 
\end{algorithmic}
\end{algorithm} 
The algorithm \ref{srkl} elucidates the learning procedure of SRKL model. The search grammar keeps playing its role as a generator of composite kernels for each depth $s$. For each kernel in the search space, the hyperparameters consist of shared hyperparamters $\theta_S$, scale factors $\sigma_1,\dotsc,\sigma_M$ , distinctive hyperparameters $\theta_1, \dotsc, \theta_M$. The optimal hyperparameters are learned based on finding the minimum negative log likelihood of data on kernels $\mathbf{K}$. The objective is to find the most common components between time series. The search grammar identifies the best shared kernel by the BIC score on shared kernel $K_S $where the likelihood part is computed by the summation of likelihood of each time series with respect to the shared kernel. 

\begin{figure}[t!]
    \centering  \includegraphics[width=0.5\textwidth]{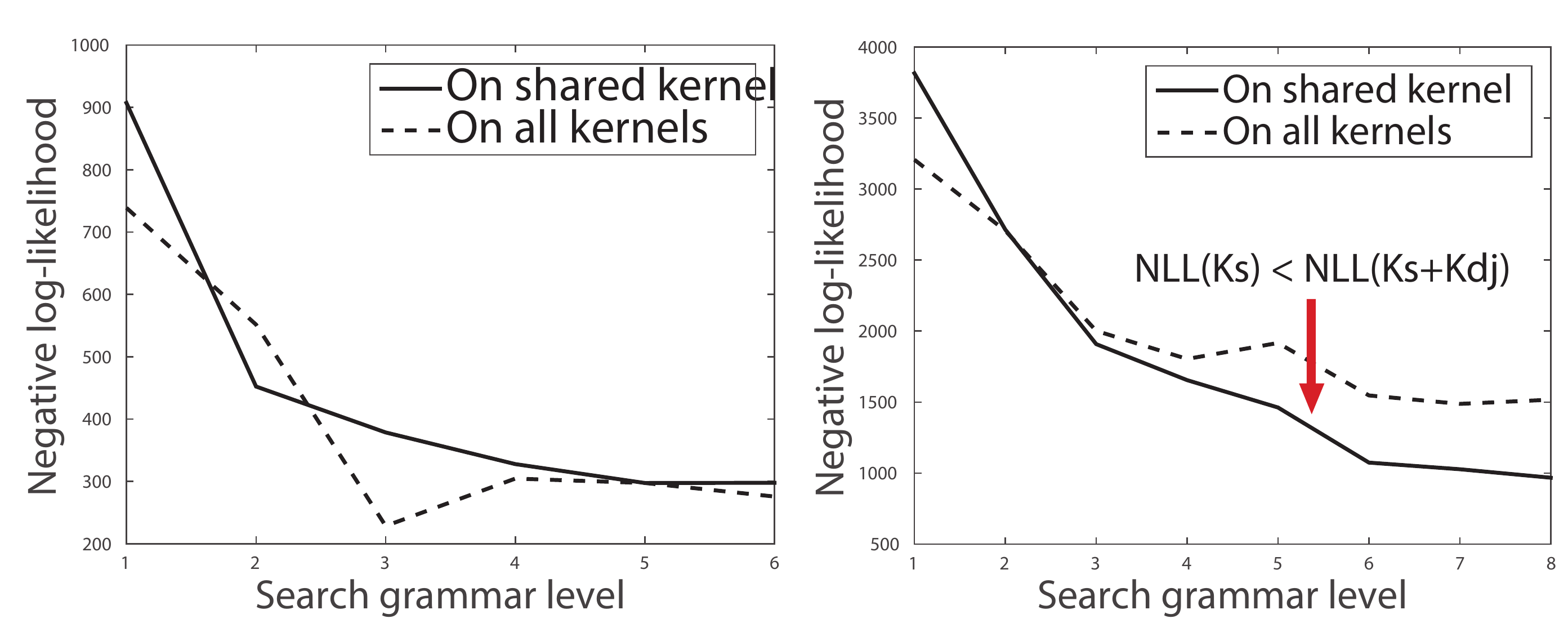}
\caption{Negative log-likelihoods (NLL) made by $K_S$ and $K_j$. (a) Non-overfitting case. The NLLs on $K_j$ and $K_S$ decrease together; (b) Overfitting starts from level 3 of search grammar. From level 3, the total NLL on $K_S$ keeps going down while the total NLL on the whole kernel $K_j$ starts increasing. $K_{d_j}$ adds more complexity to the model and worsens the $K_j$.}
\label{fig:overfit}
\end{figure}

During the learning procedure, there is a notable compromise between the shared kernel $K_S$ and distinctive kernels $K_{d_j}$. 
At the beginning (lower levels of search grammar), the shared kernel is still coarse and not expressive enough. The distinctive kernel fills the gap between the true kernel and the shared kernel. When the search grammar is getting deeper, $K_S$ becomes more complex and makes $K_{d_j}$ adapt to data not explained by $K_S$, yet. Once $K_S$ gets too complex to data, $K_{d_j}$ will be unable to make improvement on $K_j$. The overfitting  phenomena occurs when the  negative log-likelihood made by $K_j$ starts overwhelming the negative log-likelihood made by $K_S$ as shown in Figure~\ref{fig:overfit}.

\section{Related Work}
\label{sec:related}

\subsection{Learning Composite GP Kernels}
A composite GP kernel can greatly improves the performance of nonparametric regression models. Manually constructed composite models have shown to fit accurate GP models \cite{rasmussen2006gaussian,KlenskeZSH13,lloyd2014gefcom2012}. Tree-shaped composite kernels have also been used with Support Vector Machine (SVM) \cite{DiosanRP07,Bing2010}.

A weighted sum of base kernels is limited but also effective when building a composite kernel with composite kernels are given. Multiple Kernel Learning (MKL) \cite{mkl04,CandelaR05,WilsonA13} find optimal weights in polynomial time when the component kernels and parameter are pre-specified. 

Recently, an algorithm for learning a composite kernel (CKL) \cite{duvenaud4922structure}, and a system for an automatic explanation of the learned composite kernel (ABCD) \cite{lloyd2014automatic} have been proposed. Our RKL and SRKL are based on CKL and the ABCD system. However, our models expand the CKL for multiple time-series data and improves the interpretability of the individual time-series data by finding a shared relational structure.
  
\subsection{Relational GPs for Continuous Data}
Relational GPs can represent the relationships among multiple time-series data sets \cite{churelational,XuKT09,GWP}. These models are based on pre-specified base kernels instead of composite kernel representations. A straightforward extension of relational GP with a composite kernel does not work unless the time-series data sets share the same hyperparameter. Our models are more flexible to handle relational data sets by introducing GPs with both shared and distinct, non-shared hyperparameters. 

There is a large body of work attempting to represent the probabilistic knowledge formalisms in statistical relational learning (SRL) \cite{SRL}. SRL models for continuous variables \cite{WangD08,RCM,BelleIJCAI15,Choi15} and lifted inference algorithms have been proposed \cite{KimmigMG15}. Unfortunately, learning composite relational models from complex continuous data is still hard, and thus models are defined manually.

\subsection{Multiple Output GPs}
Multiple correlated data sets can be handled by multiple output GPs \cite{Wei07,Pan08,Silva08,Alvarez11,Qui16}. In principle, handling multiple output GPs is intractable because of the increased size of the full covariance function (matrix) \cite{Alvarez11}. Thus, existing work exploits efficient ways to extract a common (but simple) structure given a fixed  GP kernel. Our RKL and SRKL seek for a similar goal by finding a shared covariance kernel. However, our models  with a shared composite GP kernel and a distinctive components (scale factors and SM) are far more expressive. 

\section{Experimental Results}
\label{sec:exps}
In this section, we compare two proposed models, RKL and SRKL, with CKL, the learning algorithm of the ABCD system.  

\subsection{Data sets}
\subsubsection{Stock market}
From US stock market data, we select the 9 most valuable stocks, GE, MSFT, XOM, PFE, C, WMT, INTC, BP and AIG based on the market capitalization ranks as of  2001~\cite{topmarketcap}. The adjusted closes of stocks from 2001-05-29 to 2001-12-25 were collected from Yahoo finance~\cite{yahoofinance}. Each stock's historical adjusted close in that period consists of 129 points. Thus, the total number of points is $1161 ({=}129{\times}9)$. The collected period includes the September 11 attacks. 
After the 9/11 attacks, most of  stock values show a steep drop and gradual recovery as time goes on. We arrange this data set into three different learning settings - STOCK3, STOCK6, and STOCK9. The suffix number indicates the number of most valuable stocks.
\subsubsection{Housing Market}
US house price index data are retrieved from S\&P Dow Jones Indices~\cite{homepriceindex} as a seasonally adjusted home price index levels. We selected 6 cities: New York, Los Angeles, Chicago, Pheonix, San Diego and San Francisco based on the US city population ranks~\cite{toppoprank} and where the house price index is available.  The period is from the start of year 2004 to the end of year 2013 with monthly granularity, with total 120 points for each data set and 720 for all the data sets. There are smooth peaks around 2007 and drops until 2009 (the subprime mortgage crisis). 
Here, we have three learning settings - HOUSE2, HOUSE4, and HOUSE6 with top housing markets.
\subsubsection{Emerging Currency Market}
We collect US dollar to 4 currencies exchange rates from emerging markets;
South African Rand (ZAR), Indonesian Rupiah (IDR), Malaysian Ringgit (MYR), and Russian Rouble (RUB). The currency data from 2015-06-28 to 2015-12-30 are acquired from Yahoo Finance ~\cite{yahoofinance}, containing 132 currency values for each currency. The financial market greatly fluctuated from the end of September 2015 to the beginning of October 2015 affected by several economic events including FED's announcement about policy changes in interest rates and China's foreign exchange reserves falls. We name this data set as CURRENCY4.
\begin{figure*}[t]
\includegraphics[width=\textwidth]{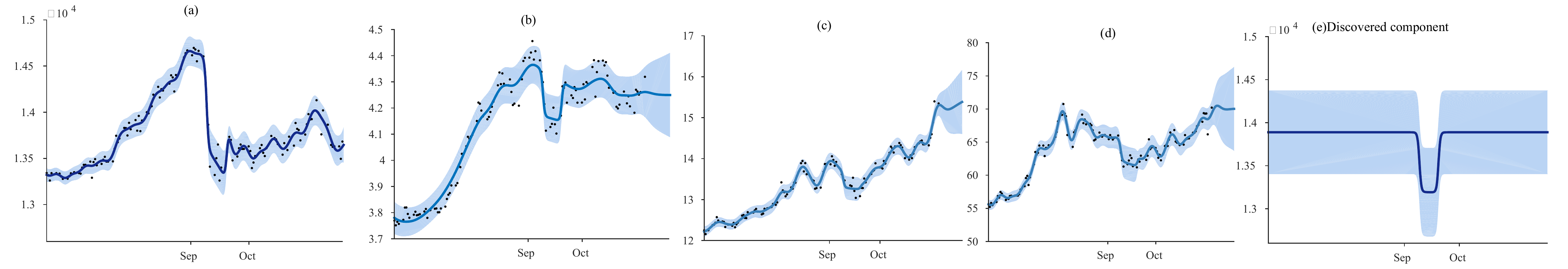}
\label{exchange_final}
\caption{A shared covariance kernel found by SRKL in the currency exchange rate data.}
\end{figure*}
\begin{table*}[ht]
\centering
\caption{Comparisons of CKL, RKL, and SRKL by NLL, BIC, and RMSE among on Us stock data, US house price index data and current exchange rate data}
\label{tab:all}
\begin{tabular}{l|ccc|ccc|ccc|}
\cline{2-10}
                                 & \multicolumn{3}{c|}{Negative log likelihood} & \multicolumn{3}{c|}{Bayesian Information Criteria} & \multicolumn{3}{c|}{Root mean square error} \\ \hline
\multicolumn{1}{|l|}{Data set}   & CKL            & RKL        & SRKL           & CKL               & RKL               & SRKL       & CKL            & RKL        & SRKL          \\ \hline
\multicolumn{1}{|l|}{STOCK3}    & 332.75         & 311.84     & \textbf{304.05} & 750.65            & \textbf{665.09}   & 1251.62    & 0.40           & 0.78       & \textbf{0.38} \\
\multicolumn{1}{|l|}{STOCK6}    & \textbf{972.00} & 1007.09    & 988.14         & 2219.71           & \textbf{2066.18}  & 3333.21    & 3.69           & 5.75       & \textbf{1.22} \\
\multicolumn{1}{|l|}{STOCK9}    & 1776.31        & 1763.96    & \textbf{1757.11} & 3985.03           & \textbf{3626.00}  & 5633.33    & 8.35           & 9.77       & \textbf{4.85} \\ \hline
\multicolumn{1}{|l|}{HOUSE2}    & \textbf{264.69} & 304.29     & 310.38         & \textbf{634.00}   & \textbf{634.76}      & 905.76     & 6.58           & \textbf{2.75} & 3.12       \\
\multicolumn{1}{|l|}{HOUSE4}    & 594.79         & \textbf{586.81} & 1249.82        & 1424.18           & \textbf{1221.88}  & 3326.94    & 5.84           & 3.66       & \textbf{2.22} \\
\multicolumn{1}{|l|}{HOUSE6}    & \textbf{849.64} & 891.09     & 1495.40        & 2100.62           & \textbf{1876.47}  & 4339.54  & 7.96           & 5.33       & \textbf{3.10} \\ \hline
\multicolumn{1}{|l|}{CURRENCY4} & \textbf{578.35} & 617.77     & 693.76         & \textbf{1165.82}  & 1291.77           & 2269.17    & 330.00         & 282.24     & \textbf{201.56} \\ \hline
\end{tabular}
\end{table*}

\subsection{Quantitative evaluations}
Table~\ref{tab:all} presents all experimental results in three different evaluation criteria; negative log-likelihood (NLL) and  BIC on training data, and root mean square error on test (extrapolation) data.
\subsubsection{Negative log likelihood and BIC}
The first two groups of columns in Table~\ref{tab:all} present how models fit data. RKL surpasses in most of data sets in term of the BIC. RKL model consistently keeps a small number of hyperparameters since it shares parameters through the data sets. With a single kernel and some scale factors, RKL express multiple sequences better than CKL in STOCK3, STOCK6, STOCK9, HOUSE4, and HOUSE6. It can be inferred that these time series are highly correlated. 

SRKL also seeks for a general, shared kernel  among time series data sets as RKL. However, the number of hyperparameters in the distinctive (spectral mixture) kernel contributes to high SRKL's BIC score. In addition, the SRKL kernels $K_j = K_S + K_{d_j}$ is more complex than other kernels. The complexity penalty $\log |K_j|/2$ ~\cite{rasmussen2006gaussian} in the negative log-likelihood term causes high negative log-likelihood. 
\subsubsection{Extrapolation performance}
We assess performance using root mean square error (RMSE) between the predicted and  actual values on the extrapolation after the end of the training period. For stock data, housing data, and currency exchange data, we respectively collected the next 14 days, 13 months, and 13 days of data.

The RMSEs of CKL, RKL, and SRKL  are presented in the third group of columns in Table~\ref{tab:all}. 
Although SRKL has been seen possessing bigger BIC scores, it outperforms on most of data sets. It conveys the general information among time series via the shared kernel. SM kernels are in charge of the distinctive information, complement favorably the shared kernel for each time series. 
RKL also outruns CKL on the housing data and the currency exchange data. 

\subsection{Qualitative Comparisons}
\Gls*{rkl} can find distinct signal components those are dominant in multiple sequences better than \gls*{ckl}. While \gls{ckl} learns a model based on only a single dataset, \gls*{rkl} can refer to multiple datasets and thus have much evidence to decide whether a certain signal is really distinct or dominant. As an example with the US stock market data, when we learned model for each data set using \gls*{ckl}, the most of the learned models could not specify the drop after the 9/11 attacks as a single component but just considered the drop as a kind of normal drift included in the smooth changes of the signal as time flows. However, in case of the \gls*{rkl}, the model could find the component that solely explains the drop by exactly specifying the time window of the sudden drop and recovery after 9/11, similar to the Figure~\ref{fig:compare}. 

In currency exchange rate data, SRKL also finds  a qualitatively important compositional kernel shortly written as CW(SE + CW(WN + SE, WN), CONST). The second change-window kernel presents a time period from mid September 2015 to mid October 2015. This reveals the big changes in financial markets influenced by fiscal events, FED's announcement about policy changes in interest rates. ABCD captures a change-point on only one currency for Indonesian Rupiah. The other results form ABCD do not show this change.

\section{Conclusion}
\label{sec:conclusions}

We proposed a nonparametric Bayesian framework that finds shared structure throughout the multiple sets of data. The resulting Relational Kernel Learning (RKL) method provide a way of finding a shared kernel function that can describe multiple data with better BIC. We applied this model to a synthetic data and several real world data, including US stock data, US house price index data and currency exchange rate data, to validate our approach. 

\bibliographystyle{icml2016}
\bibliography{RKL}

\clearpage

\appendix
\onecolumn
\section{Appendix}

\subsection{Components Extracted from the Stock Market Data}

\begin{figure}[h!]
\centering
\includegraphics[width=0.70\textwidth]{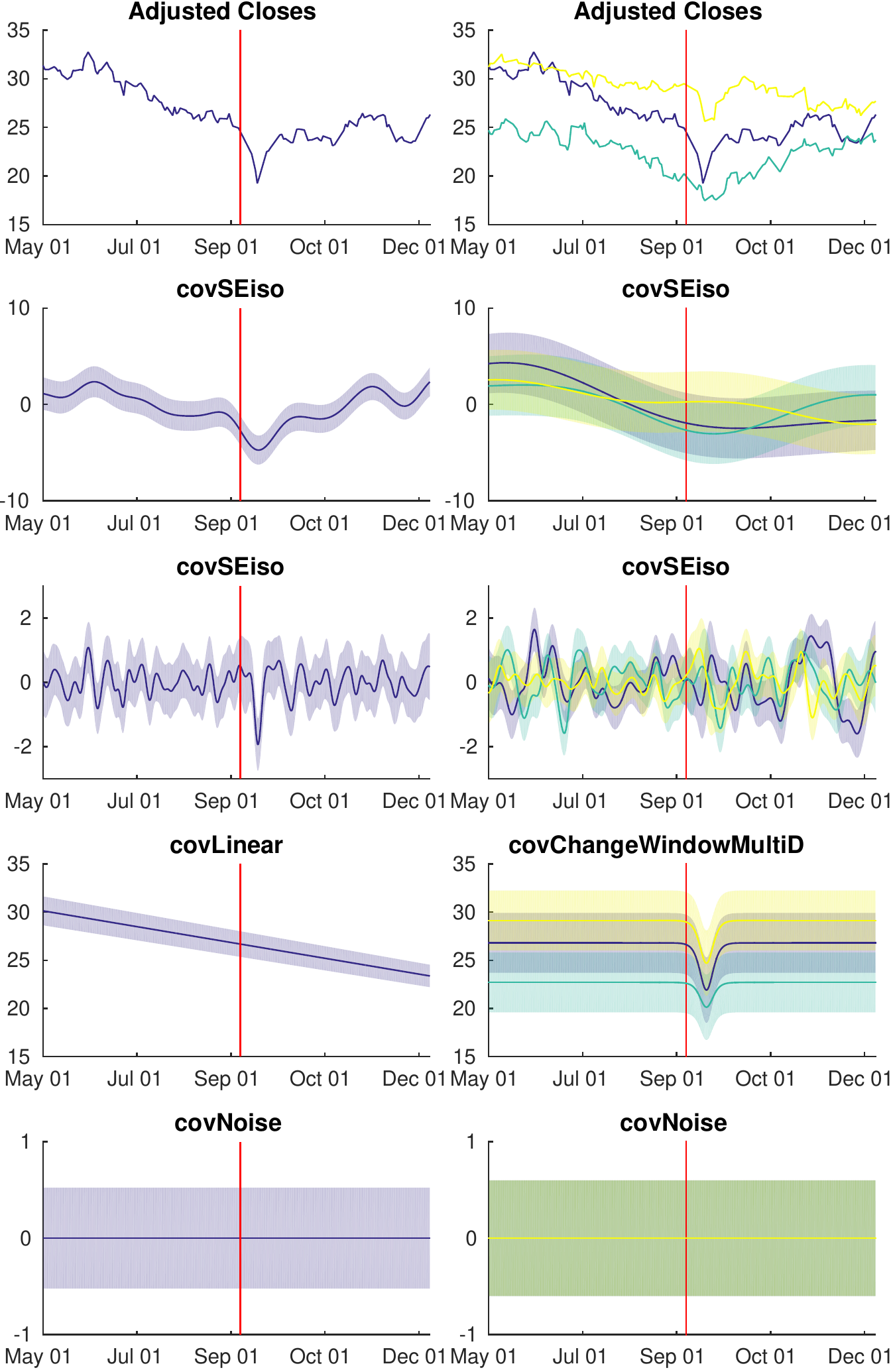}
\caption{Figure that shows decomposition of adjusted closes of stock values. $x$ axis is time and $y$ axis is stock values. Red vertical line is at `2001/09/11' which is the day of 911. The third component of right side, `covChangeWindowMultiD' captures sudden drop of stock values after the 911.}
\label{fig:multi}
\end{figure}

\subsection{Learning Hyperparameters in Synthetic Data}

\begin{figure}
\centering
\includegraphics[width=0.5\textwidth]{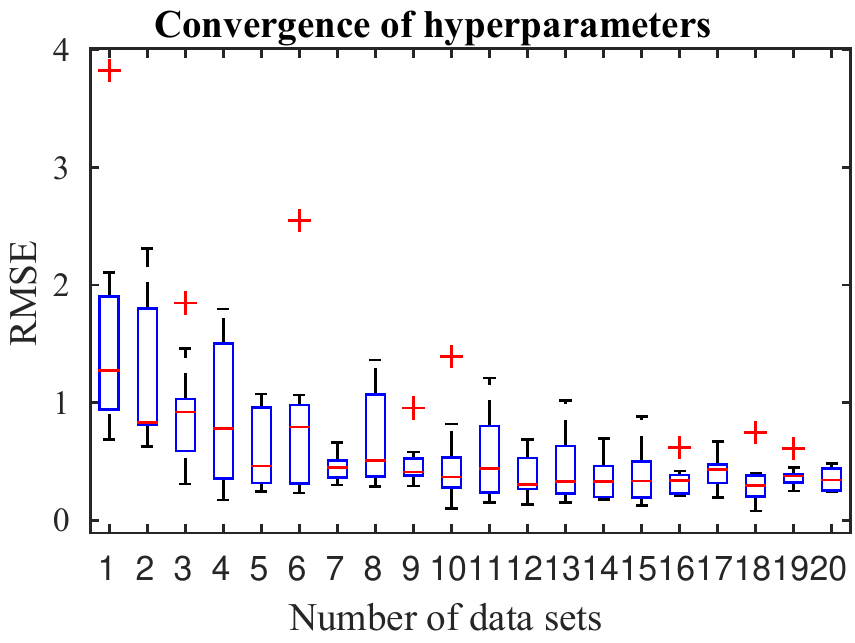}
\caption{The convergence of error between the optimized hyperparameter and original true parameter that was used when generating data. The horizontal axis denotes the number of data sets used in hyperparameter optimization. The vertical axis shows the value of root mean squared error between optimized and true hyperparameters. Each boxplot shows the distribution of errors for each step.}
\label{fig:converge}
\end{figure}

First we have a covariance kernel function $k$ that is already learned from the original \acrshort*{abcd} algorithm. 
We generate 20 data sets from a GP prior with kernel $k$. Each data set has a different number of data points. After sampling the data, we try to optimize the hyperparameters of $k$ for $1,2,\dotsc,20$ data sets chosen out of 20 sampled data sets, with the original hyperparameters as a starting point. We do this optimization 10 times for each number of data sets, from 1 to 20, with different combination of choice. 

We calculate the root mean squared error (norm distance) between the true hyperparameter and the optimized hyperparameter vector. Figure~\ref{fig:converge} shows the result. The values along the horizontal axis means the number of data sets which are used in the optimization. Each boxplot shows the distribution of the error with 10 different results for each number of data sets used. It shows that as we increase the number of data sets in parameter optimization, the error between the optimized and the true hyperparameter decreases.

\subsection{Experiment table with 6 stocks and 6 cities}

\begin{table}[t]
\centering
\begin{tabular}{|l|l|r|r|r|}
\hline
\multicolumn{2}{|c|}{Model-Stock} &NLL	&P	&BIC \\
\hline
\multirow{7}{*}{CKL} & GE & 116.36  & 7 & 266.75 \\
& MSFT & 111.17  & 6 & 251.51 \\
& XOM & 82.24 & 9 & 208.23 \\
& PFE & 62.91  & 12 & 184.14 \\
& C & 424.23  & 10 & 897.07 \\
& WMT & 149.96  & 5 & 324.23 \\ \cline{2-5}
& Total	& \textbf{946.89}  & 49 & 2219.71 \\
\hline
\multicolumn{2}{|c|}{RKL-Total} & 1007.09  & \textbf{18} & \textbf{2066.18} \\
\hline
\multicolumn{2}{|c|}{SRKL-Total} & 988.15  & 204 & 3333.21\\
\hline
\end{tabular}
\caption{The experimental comparisons of GP models with individual CKL, RKL and SRKL in the stock data set (top 6 US stocks in 2001). For each column, NLL means the negative log likelihood, P is the number of hyperparameters and BIC is the Bayesian information criterion.
CKL is slightly better in a simple aggregation of NLL only. However, our RKL outperforms CKL in the BIC due to a significantly lower 21 parameters (P) compared to 49 parameters in CKL.}\label{tab:stock}
\end{table}

\begin{table*}[t]
\centering
\begin{tabular}{|l|r|r|r|r|r|r|r|r|r|}
\hline
&  & \multicolumn{2}{c|}{RKL} & \multicolumn{2}{c|}{SRKL} & \multicolumn{2}{c|}{CKL} \\
\hline
Stocks & N & P & BIC & P& BIC& P & BIC \\
\hline
Top 3 & 387 & \textbf{16} & \textbf{665.09} &108&1251.62&22 & 750.65 \\
Top 6  & 774 & \textbf{18} & \textbf{2066.18} & 204 & 3333.21& 49 & 2219.71 \\
Top 9 & 1161 & \textbf{32} & \textbf{3626.00} &300& 5633.33 &  73 & 3985.03\\
\hline
\end{tabular}
\caption{The BIC of CKL, RKL and SRKL in the stock data set. `Top 3', `Top 6' and `Top 9' stocks were selected by their market capitalization ranks in 2011. As shown in Table~\ref{tab:stock}, RKL requires fewer parameters than CKL. RKL models trained with 3 stocks and 6 stocks show better performance than individually optimized CKL models. When 9 stocks are considered, the individual CKL models show better performance than the single (shared) RKL model.}
\label{tab:stock369}
\end{table*}

Table~\ref{tab:stock} compares the fitness of the CKL, RKL and SRKL models. The results with line heading RKL are from our model. The others are from the original ABCD. The results with line heading Total is calculated by considering the whole GP priors that are learned for each data set as a single model. So NLL and P values of individual models are added up and the total BIC of the CKL model as a single model was calculated using those values. In terms of negative log likelihood, applying \acrshort*{abcd} individually gives better results. However if we compare the BIC, our model achieves better results as our model has a reduced number of free parameters by sharing them through the data sets. 

Table~\ref{tab:stock369} compares the fitness of the models with different numbers of data sets used in training. As the number of data sets increases, the number of parameters needed also increases. However RKL shows less need of parameters through the whole setup since our model shares parameters though the data. RKL also performs well in terms of the BIC for TOP3 and TOP6. However as the number of data sets increases, the performance of the original CKL model surpasses our model, at TOP9. This is because our model fits the general structure but not individual specific ones. As the number of data sets increases, the learned structure cannot fully explain the individual specific patterns. And this accumulated errors in fitness, which is the NLL, offsets the advantages of the BIC which is from the reduced number of parameters.

\begin{table}[t!]
\centering
\begin{tabular}{|l|l|r|r|r|}
\hline
\multicolumn{2}{|c|}{Model-City} &NLL	&P	&BIC \\
\hline
\multirow{7}{*}{CKL} & New York & 120.13  & 10 & 288.13 \\
& Los Angeles & 162.94  & 9 & 368.96 \\
& Chicago & 134.73 & 13 & 331.69\\
& Pheonix & 101.76  & 11 & 256.17 \\
& San Diego & 155.64  & 12 & 368.73 \\
& San Francisco & 174.45  & 6 & 377.63 \\ \cline{2-5}
& Total	& \textbf{849.64}  & 61 & 2100.62 \\
\hline
\multicolumn{2}{|c|}{RKL-Total} & 891.09  & \textbf{33} & \textbf{1972.58} \\
\hline
\multicolumn{2}{|c|}{SRKL-Total} & 1495.40  & $205$ & 3707.94\\
\hline
\end{tabular}
\caption{A comparison of BIC between  individual \acrshort*{ckl} \acrshort*{rkl} and SRKL, with house data. For each column, NLL means negative log likelihood, P is the number of hyperparameters and BIC is the Bayesian information criterion. For each row, RKL is the result from our model. From New York to San Francisco, those results are from individual cities. Total is summation of those individual results.}\label{tab:house}
\end{table}

\begin{table*}
\centering
\begin{tabular}{|l|r|r|r|r|r|r|r|r|r|}
\hline
&  & \multicolumn{2}{c|}{RKL} & \multicolumn{2}{c|}{SRKL} & \multicolumn{2}{c|}{CKL} \\
\hline
SET & N & P & BIC & P & BIC & P & BIC\\
\hline
Top 2 cities & 240 & \textbf{11} & 634.76 & 52& 905.76& 20 & \textbf{634.00} \\
Top 4 cities & 480 & \textbf{18} & \textbf{1221.88} & 134 & 3326.94 & 38 & 1424.18 \\
Top 6 cities & 720 & \textbf{33} & \textbf{1876.47} & 109 & 4339.54 & 61 & 2100.62 \\
\hline
\end{tabular}
\caption{The BIC of CKL, RKL and SRKL in the housing market data set. `Top 2', `Top 4' and `Top 6' US cities were selected in terms of their city population rank. The BICs of the RKL models are similar or better than the BICs of individually trained CKL models.}
\label{tab:house246}
\end{table*}

Table~\ref{tab:house} shows the fitness of models between our model and the original ABCD model. Similar to the stock market data case, the first line is our model and the others are from the original ABCD. Our model shows better results, reduced number of free parameters and smaller BIC compared to Total case. However the original model still shows better fitness if we only consider NLL. From this experiment we can again confirm that the major contribution of the smaller BIC comes from the reduce number of parameters.

Table~\ref{tab:house246} compares results between our model and the ABCD for different number of data sets. Our model shows reduced number of parameters over all different sets of data since our model shares kernel parameters for multiple data sets. The original model shows better results in terms of BIC for the top 2 indices case. However other than that, our model shows better results in terms of both number of parameters and BIC.\\

\end{document}